# Adaptive color transfer from images to terrain visualizations

Mingguang Wu[a,b,c,d] Yanjie Sun[a,b]* and Shangjing Jiang[a,b]

[a]*Key Laboratory of Virtual Geographic Environment of Ministry of Education, Nanjing Normal University, Nanjing, Jiangsu, China;* [b]*College of Geographic Sciences, Nanjing Normal University, Nanjing, Jiangsu, China;* [c]*State Key Laboratory Cultivation Base of Geographic Environment Evolution (Jiangsu Province), Nanjing, Jiangsu, China;* [d]*Jiangsu Center for Collaborative Innovation in Geographical Information Resource Development and Application, Nanjing, Jiangsu, China*

[Corresponding author] Yanjie Sun, E-mail: 201301022@njnu.edu.cn

[E-mail addresses] wmg@njnu.edu.cn (Mingguang Wu), 201301022@njnu.edu.cn (Yanjie Sun), 211301018@njnu.edu.cn (Shangjing Jiang)

# Adaptive color transfer from images to terrain visualizations

**Abstract:** Terrain mapping is not only dedicated to communicating how high or how steep a landscape is but can also help to narrate how we feel about a place. However, crafting effective and expressive hypsometric tints is challenging for both nonexperts and experts. In this paper, we present a two-step image-to-terrain color transfer method that can transfer color from arbitrary images to diverse terrain models. First, we present a new image color organization method that organizes discrete, irregular image colors into a continuous, regular color grid that facilitates a series of color operations, such as local and global searching, categorical color selection and sequential color interpolation. Second, we quantify a series of subjective concerns about elevation color crafting, such as "the lower, the higher" principle, color conventions, and aerial perspectives. We also define color similarity between image and terrain visualization with aesthetic quality. We then mathematically formulate image-to-terrain color transfer as a dual-objective optimization problem and offer a heuristic searching method to solve the problem. Finally, we compare elevation tints from our method with a standard color scheme on four test terrains. The evaluations show that the hypsometric tints from the proposed method can work as effectively as the standard scheme and that our tints are more visually favorable. We also showcase that our method can transfer emotion from image to terrain visualization.

# 1. Introduction

In this paper, we present an automatic method to transfer color from arbitrary images to terrain visualizations, which can facilitate making expressive terrain maps for both nonexperts and experts. With centuries of experience and innovation, cartographers have developed a series of methods to present terrain, such as profiles, form lines, contour, hillshade, and hachure (Brewer and Marlow 1993, Häberling and Hurni 2002, Imhof 1982, Kennelly 2009). Among them, hypsometric tints, also called elevation colors, are critical to make effective and vivid terrain visualizations. By matching colors with elevations, elevation zones can be identified; leveraging color stereoscopic effect, three-dimensional appearance can be enhanced (often with contour and hillshade) (Eyton 1990). Hypsometric tints also play an important role in making terrain maps visually favorable.

Both naturalism and symbolism should be considered when crafting expressive hypsometric tints. Natural images, such as aerial photography, imitate the natural color of terrain, carrying an impression of reality, but the natural color of terrain may vary with weather conditions, seasons and/or observation times (e.g., sun rise or sun set). Generally, there is not a constant natural color for a patch, let alone for an elevation zone. Moreover, natural color may not always be effective for distinguishing. For example, different elevation zones may fall into the same color, introducing confusion when using color differences to distinguish elevation. On the other hand, cartographers often select representative colors to symbolize elevation zones, such as green for lowland and white for highland. As terrain is geographically variable across space, which color is representative? It may depend on location and observers.

The diversity of both terrains and observers should be considered to make expressive terrain maps. Contemporarily, terrain maps no longer exist only in atlases or on wall maps, and they are also favored by nonexperts to narrate an experience using a landscape, e.g., making a story map of a wonderful road trip. In this case, not only elevation and terrain features (such as landform, slope, etc., but also aesthetic and emotional aspects of the landscape should be addressed. Casey (2005) stated that "*such mapping concerns the way one experiences certain parts of the known world: the issue is no longer how to get there or just where there is in world-space, but how it feels to be there, with/in that very place or region*". Mapping experience is not a new concept. In practice, an observer's point of view (looking up from low altitude or down from high altitude or at ground level) and distance from the landscape impact its visual impression. Accordingly, a series of principles, such as "the higher, the lighter", "the richer in color, the nearer", and aerial perspective, are suggested (Imhof 1982). However, individuality, such as preference and emotional response to a landscape, is not considered when crafting elevation colors.

In this study, we focus on transferring color from images to terrain maps to make effective and expressive hypsometric tints. Color transfer refers to transferring the visual impact of the color of one image to another (Gatys et al. 2016). Because designing expressive colors from scratch is difficult, a common approach is to find an inspiration, such as a master's painting or an impressive photograph. Considerable work has been done on transferring color from image-to-image in the field of computer graphics, such as neural network-based color methods (Isola et al. 2017, Li et al. 2018) (which will be further discussed in Section 2.2). It also draws the attention of many cartographers. Wu et al. (2021) present an adaptive method to transfer color from an image to a map, however, it is reference map-oriented. Computational image-to-terrain color transfer is not yet available.

To fill this gap, we present an adaptive method that can transfer color from arbitrary images to terrain visualizations. We present a new method to extract and organize dominate colors from images. We also quantify a series of considerations for making elevation colors, such as "the higher, the lighter" and aerial perspective, and then formulate the problem of color transfer as a dual-objective optimization problem. We implement our method with four terrain samples. The developed approach is available as an open-source package distributed under the MIT License at https://doi.org/10.5281/zenodo.4727805 for reproduction and extension.

In the following, we summarize related work on crafting elevation colors and color transfer in Section 2. In Section 3, we introduce an overview of the proposed method and then present the details of how to formulize and solve the problem. We evaluate our method in the fourth section by comparing its outputs with a standard color scheme. We discuss the advantages and limitations of the proposed method in the fifth section. Finally, we conclude the paper with a summary of the key contributions and outline future studies in Section 6.

## 2. Related work

Considerable work has been carried out on crafting elevation colors. Here, we summarize major principles on crafting elevation colors, and then we review available computational color transfer methods.

## 2.1 Principles of hypsometric tints

**Continuity of elevation.** The natural continuity of elevation requires a corresponding continuity in hypsometric tints to be applied. Two opposite principles exist: "the higher, the darker" and "the higher, the lighter" (Imhof 1982). "The higher, the lighter" can be interpreted as the natural result of diffuse lighting overhead (the sun). The closer an object is to the light source

(the higher the altitude), the lighter it becomes (Imhof 1982). Themes for low-layer zones (e.g., building and transportation) are often conventionally symbolized with saturated color as they will become less readable if applying dark tints as background, or even worse when hypsometric tints are combined with shaded relief. In contrast, "the higher, the darker" scheme keeps the lowland lighter, but mountainous regions will become darker overall and therefore lack a three-dimensional appearance (Patterson and Jenny 2011). In practice, these two principles may be weighted or combined for specific mapping contexts.

**Color conventions.** Hypsometric colors can be assigned naturally according to the real landscape. However, the reality for an elevation zone may vary across space and over time. Instead, colors can be set up in a symbolic way, such as green for lowlands, yellow for mid-elevations, and white for highlands. Color conventions can be interpreted as a resonance between a color and a semantic meaning, such as greenish colors for fertile plains and white for snow (Imhof 1982, Patterson and Jenny 2011). Color conventions are now widely accepted, especially in standardized topological maps.

However, conventional colors, as a one-size-fits-all strategy, may introduce misleading results (Jolivet and Renard 2009). For instance, a green lowland color in a dry desert and a red color in highlands rich in forests may confuse map readers (Imhof 1982). Patterson and Jenny (2011) designed environment-aware hypsometric tints, in which environmental factors, such as polar, humid, and arid conditions, are distinguished, anticipating computational methods that can weight and combine naturalism and symbolism.

**Aerial perspective.** Due to dust and droplets in the atmosphere, light can be scattered, resulting in atmospheric haze over the landscape, which is called the aerial perspective (Imhof

1982). Correspondingly, visual contrast may become lower with increasing distance. For hypsometric tints that consider the aerial perspective, closer elevation zones should be colorized with higher brightness and color contrast (Jenny 2001).

Many efforts are now available to make aerial perspectives with reliefs, such as luminance adjustments (Brassel 1974, Jenny and Patterson 2021), elevation masks (Patterson 2001), and neural networks (Jenny et al. 2020).

**Aesthetics.** Visual pleasure is constructive to an object's value as a creation but not in a rational and functional way in terms of practical use (e.g., identification tasks). Imhof (1982) observed that "*the greatest clarity, the greatest power of expression, balance and simplicity are concurrent with beauty*". Ideally, like other types of visual works, terrain maps should be aesthetically colorized for reading and be visually pleasing.

Elevation colors therefore should be compatible. There are well-accepted principles to craft compatible color schemes, such as rule-based (e.g., Munsell's four rules on color harmony (CLELANDT 1937)), distance-based (Moon and Spencer 1944), template-based (e.g., Matsuda's eight templates on color wheel (Matsuda 1995)), and learning-based models (O'Donovan et al. 2011). Color aesthetics are also dominated by preference, which may vary by individual, age, gender, and/or nation and over time.

**Emotion**. While colors can encode semantic meaning, colors can also carry affective connotations. Map colors can be operationalized to evoke specific emotions when reading the map (Montello et al. 2018). More specifically, while emotionally consistent colors amplify the impact of map themes, emotional inconsistencies can confuse map readers (Anderson 2018). For elevation colors, feelings about landscape are partially considered, such as warm colors for high altitude and cold colors for lowlands, coinciding with the principle of "the closer, the warmer" with an assumption of sunlight overhead. However, emotions when facing a landscape, such as joy and excitement, are not considered.

## 2.2 Color transfer

Color transfer refers to the process of imposing the color characteristics of one image onto another (Reinhard et al. 2001). Existing image-to-image color transfer methods can be grouped into three categories: statistical-based (Neumann and Neumann 2005, Reinhard et al. 2001, Senanayake and Alexander 2007), content-based (Tsai et al. 2016, Wu et al. 2013), and neural network-based (Gatys et al. 2016, Isola et al. 2017, Zhu et al. 2017) methods.

In the first group, an image's color appearance is characterized as a series of statistical color factors, such as the mean and deviation (Reinhard et al. 2001) and histogram (Neumann and Neumann 2005, Senanayake and Alexander 2007). Then, color transfer is conducted to statistically match those factors between images. These global statistical factors may be too general for a given image, such that local characteristics (such as contrast and topology) cannot be preserved well. Since image content is not explicitly modeled, mismatching on semantics may be largely introduced.

In the second group, image content is detected and then matched. Greenfield and House (2005) segment an image into regions according to the color similarity among pixels. Representative color with proportion of each segment can then be detected. By establishing representative color matching between the reference and target images, chromatic content can be preserved. Generally, the Greenfield and House method (2005) is limited in nonphotorealistic images. Wu et al. (2013) segmented images into regions using scene analysis (such as sky, building, vertical) according to the clarity of the image content. Matching those regions, color transfer with visual coherence can be achieved. Additional constants, such as saliency, can be further incorporated to make more sophisticated color matching (Frigo et al. 2014). While content is addressed well in this group, discussion on style of reference (e.g., tone and aesthetic quality) is lacking.

Both style and content are considered in neural network-based methods. Convolutional neural networks (CNNs) are introduced to extract image characteristics and conduct style transfer (including color) between images (Gatys et al. 2016). In this method, a deep neural network with multiple hidden layers is trained to represent the hierarchy information of an image, from low-level pixel values to high-level objects and their arrangement, called content. In addition to content, texture information can also be obtained by using a Gram matrix, which indicates the feature correlation of multiple layers, called style. The content of the target image and style of the reference image can then be weighed and synthesized to make a restyled image.

While learning an adaptive loss function to balance diverse styles and image content is challenging for CNNs, Isola et al. (2017) use a generative adversarial network (GAN) to recolor an image. A GAN introduces a classifier to guide loss learning (such as an output image that cannot be blurry) to adapt to data. A series of tools are now available to conduct image-to-image style transfer, such as *pix2pix* for paired trained data (Isola et al. 2017) and CycleGAN for unpaired training data (Zhu et al. 2017).

There are few existing studies on color or style transfer in terrain mapping. Bratkova et al. (2009) proposed an algorithm for automatically generating panoramic maps, in which two artist-cartographer color styles (e.g., base colors and brushstroke colors) are analyzed and algorithmically applied to a terrain. Jenny and Jenny (2013) proposed an example-based texture synthesis method to transfer the appearance of hand-painted panorama to a digital panorama map. Recently, Jenny et al. (2020) trained a deep neural network using manual relief shading images as samples. The result can transfer trained Swiss style to digital elevation models with high quality. However, elevation colors have not yet been considered. For cartographic color transfer, Wu et al. (2021) reported an algorithm that can extract color palates from paintings and photos and then assign them to discrete cartographic features, such as roads, rivers and parcels. It can also compose a sequential color scheme, but how to incorporate the terrain model and the aforementioned concerns are not discussed.

To date, there is no method that can transfer color from arbitrary images to terrain maps with consideration of the aforementioned principles. Systematically operationalizing the above concerns remains challenging.

## 3. Method

## 3.1 Overview

We aim to generate hypsometric tints from arbitrary images for diverse terrain models. Artists and photographers may have already captured expressive visual experience on diverse landscapes. We believe that paintings and photos of landscapes can provide inspiration to craft high-quality elevation colors. In this study, we extract expressive colors from any image type and then apply them to terrains. Elevation colors can be applied in hillshades, they can be used to color illuminated contours and as triangular irregular networks (TINs). In this study, three types of terrain representations are considered: hillshades, TINs and illuminated contours.

We also aim to computationally automate color transfer for both nonexperts and experts. Nonexperts may not have enough knowledge or skill to address the aforementioned concerns; it is also time- and effort-consuming for experts to weigh and compromise the above concerns. We provide an automatic color transfer method for both: nonexperts can pick up the resulting colors directly without any prior knowledge on terrain mapping, and expects can customize the process to generate a series of candidates.

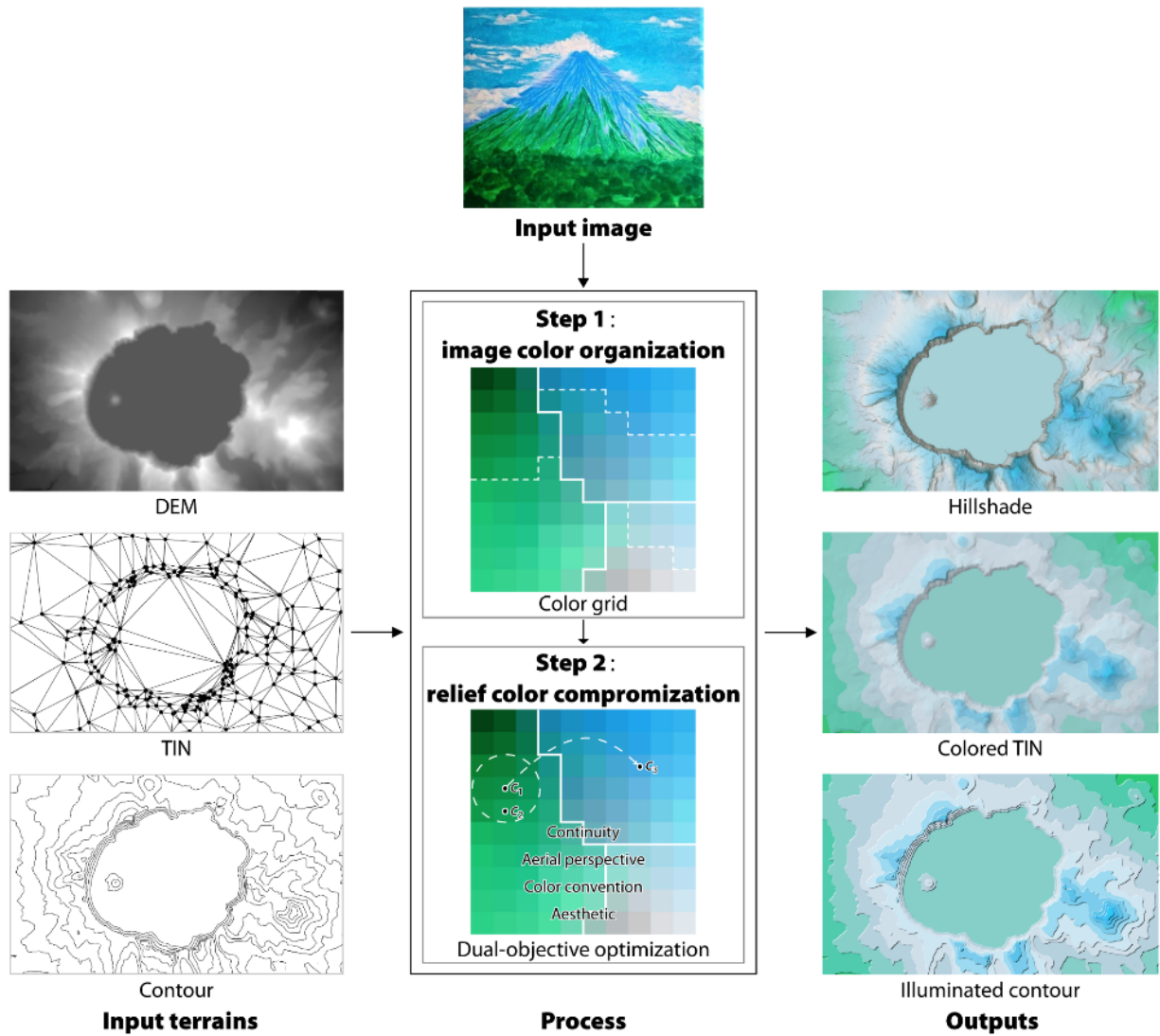

Figure 1. Diagram of our two-step image-to-terrain color transfer method.

Figure 1 shows our two-step method. First, for a given image, we extract salient colors. We then organize the extracted colors into a hierarchical color grid for further color operations. Second, we search the color grid according to subjective and aesthetic concerns and output satisfying colors in graded or continuous mode. As mentioned in Section 2.1, affective response is also a major concern when mapping landscape experience. As emotion is sensitive to diverse factors, such as content and texture, we do not quantify emotional facts in our model. However, we explicitly quantify visual similarity between the reference image and the resulting terrain maps. We aim to achieve higher color similarity, hoping to transfer emotion from the reference image to terrain maps as well, which will be evaluated in Section 4.2. In the following, we discuss the two steps in detail.

## 3.2 Image color organization

Embodied as a discrete and irregular color point cloud, an image's color gamut is extremely computationally expensive to explore for finding the desired color combination. An image's color gamut may also include noise, such as a color that appears in the image as a tiny blob or trivial stroke. Hence, we extract salient colors from the image and then organize them into a continuous and regular space for further effective color operations (e.g., searching and interpolation). We further identify dominant colors in the color grid from further color selection. Our image color organization consists of three consequent steps: 1) salient color extraction, 2) structurization and 3) dominant color identification.

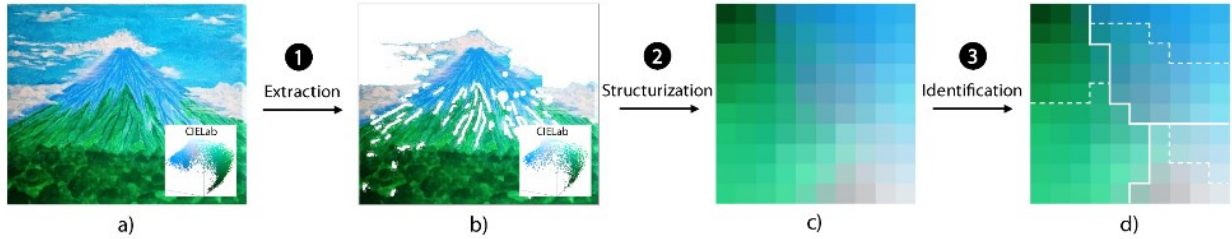

Figure 2. Three steps of image color organization: a) a thumbnail of a reference image from the internet; b) step 1: extracting salient colors from the image; c) step 2: structuring two-dimensional color grid; and d) step 3: identifying dominant colors.

**Salient color extraction**. We consider two types of elevation colors: a graded color scheme, which consists of distinct colors with noticeable color intervals, and a continuous color scheme with smooth color transitions. The second can be considered as a derivation of the first using color interpolation. We therefore extract salient colors and then interpolate them if needed. To transfer color to maps, Wu et al. (2021) separate images into blobs, edges, patches and backgrounds according to the changing intensity and visual saliency. While the features in a map are discrete and abruptly change, the terrain is continuous. Furthermore, a landscape painting or photo may not embody clear figure-ground separation (e.g., the areal perspective effort). We therefore do not separate figures from the ground, but extract color patches with relatively high saliency. A series of methods are available to measure an image's saliency. Among them, Zhu et al. (2014)'s method is adaptive to diverse image content. We therefore use it to extract visually continuous color blocks, as shown in Figure 2b.

While travel colors are reduced by selecting only salient colors, the remaining colors still remain visually indiscriminating with unnoticeable color distances. We pick up $M$ visually distinguishable colors from the extracted salient colors using $k$-means; the color proportion of color $i$ ($0 < i < M$) is accumulated through all colors in the $i$-th cluster.

**Structurization.** As shown in Figure 2b, similar to an image's color gamut, salient colors distributed across a three-dimensional space are extracted, which is challenging for both brute-force and heuristic searching. We use the self-organizing map (SOM) algorithm (Kohonen 1990) to organize them into a two-dimensional space (hereafter called the color grid). The SOM algorithm is a two-layer unsupervised neural network algorithm. Applying competitive learning, it can produce a nonlinear mapping from the $D$-dimensional space to an $F$-dimensional ($D$ is often lower than $F$) grid of weight nodes. Compared with the general dimensional-reduction method, the SOM algorithm can preserve the topological structure of the input samples.

We train the SOM network by iteratively adjusting the node weight according to its distance to the extracted colors in CIELab color space; we use the winner-takes-all strategy to update the weights: a node's weight was assigned to the color value of the closest input color. Then, a $w \times w$ color grid can be obtained, in which all colors node weights at each cell come from the above distinguishable colors. In this manner, discrete distinguishable colors in three-dimensional CIELab are organized into a continuous two-dimensional grid. Benefitting from the SOM algorithm, topology (such as adjacency) among distinguishable colors is persevered well in the color grid.

**Dominant color identification**. In addition to topology, color category and tone also needed to be compromised among the aforementioned concerns. For example, visual perception may be guided by color category; affective response is much more likely to be associated with color tones. We further segment the color grid into regions. The central point in each region is determined as a dominate color.

Here, we choose the hierarchical clustering algorithm to segment a color grid with dominant color identification. To avoid the parameter sensitivity caused by directly specifying the count of clusters, we use a just noticeable color distance (JNCD) threshold to conduct bottom-up hierarchical clustering. If the count of clustered regions is much more than the count of the needed color category, then we further cluster those regions by using two times the JNCD threshold. Clustering can be continued with increased clustering threshold until we obtain the proper number of dominant colors. As shown in Figure 2d, after clustering twice, we obtain three dominant colors in a two-layer color grid.

Our image color organization is able to facilitate both brute-force and heuristic color searching. Local color adjustment can be executed within a region. For example, if $c_1$ is assigned to represent lowland, according to color conventions (e.g., green for lowland, blue for the sea), $c_1$ generally matches the elevation zone in terms of green hue, but it is too dark according to "the lower, the lighter" principle. Then, a similar but lighter color, $c_2$, can be found around $c_1$. If $c_1$ is assigned to represent the sea, it breaks color conventions. Local adjustment cannot meet the need at all because all around colors are greenish. In this case, global exploration is needed, which shifts to a distant color region for searching. Jumping to another candidate color, such as $c_4$, color conventions can be weighed again. In this manner, our color grid can facilitate both local and global searching.

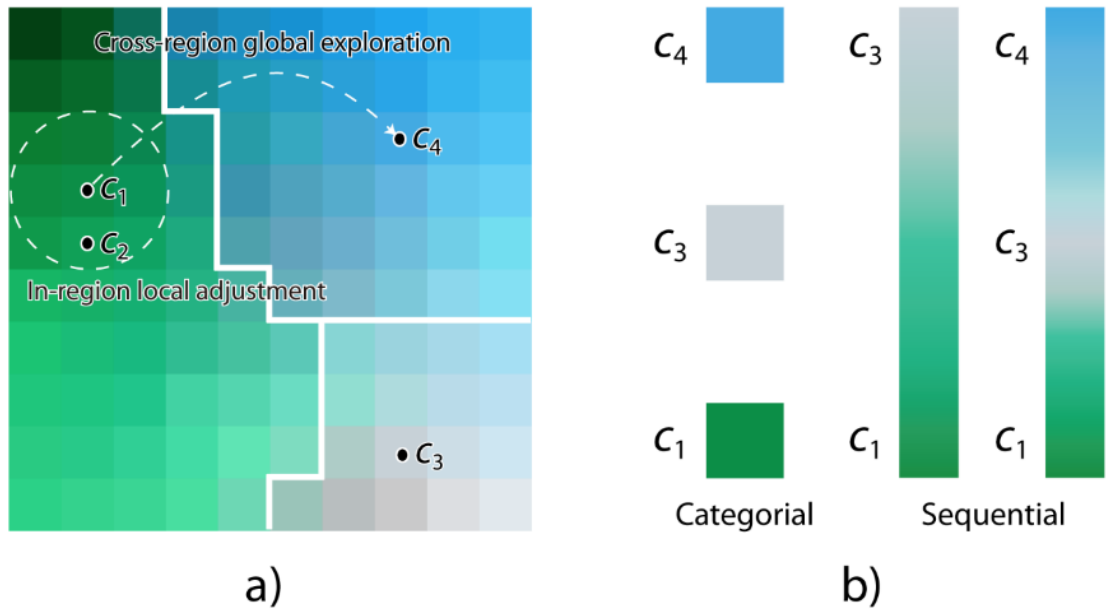

Figure 3. Color operation by using the suggested color grid: a) local adjustment and global exploration and b) generation of categorical and sequential color scheme.

Our image color organization is also able to facilitate making two primitive types of elevation tints: categorical and sequential. The categorical scheme embodies hue differences, which help to distinguish different elevation zones; the categorical scheme can be picked up from different grid regions. For example, a three-color scheme can be found by sampling three distant regions (see Figure 3b). Encoding smooth changes on evaluations, a sequential color scheme can be generated by interpolating two colors from two different grid regions. For example, as shown in Figure 3b, a green–gray sequential scheme can be sampled by using $c_1$ and $c_3$, and a green–gray-blue scheme can also be made by using continuous interpolation among $c_1$, $c_3$ and $c_4$.

In general, existing color palate extracting methods are primarily dedicated to color selection by hand, and our color grid is designed to facilitate color computation. In the following section, we further introduce how to make expressive elevation colors with the compromise of the aforementioned concerns with the color grid.

## 3.3 Relief color compromise

Similarly, making the results look like the references is a widely used evaluation criterion among current color transfer methods (e.g., histogram and semantic matching and loss in CNNs). Gatys et al. (2016) noted that the criteria of similarity are "neither mathematic precise nor university agreed upon". For elevation colors, Imhof (1982) noted: "*Objective considerations alone have not always been the deciding factors. Tradition, partiality and whim, preconceived opinions, aesthetic sensitivity or barbarity of taste often play leading roles in the selection of colors.*" We therefore refine evaluation criteria as follows. We group all concerns mentioned in Section 2.1 into two: first, subjective concerns, including continuity, aerial perspective and color conventions; and second, aesthetic concern merges into visual similarity, namely similarity with aesthetic quality. We do not explicitly encode affective concern. However, we hope to provoke similar emotions by making a visually closed terrain map with a reference. We discuss these two criteria in detail.

### 3.3.1 Operationalization of subjective concerns

**Continuity**. Both luminance and saturation are effective depth cues for the human visual system (Guibal and Dresp 2004, Egusa 1983). Here, we use the goodness of fit of both luminance and saturation to measure the continuous variation of hypsometric tints $C$, denoted as $f_g(C)$, as follows:

$$\begin{cases} f_g(C) = R_k^2(C.L) * R_k^2(C.ch_{ab}^*) * f_L(t) \\ f_L(t) = \begin{cases} 1 & if\ a_{k,C.L} * t > 0 \\ 0 & otherwise \end{cases} \quad k = 1,\ 2,\ 3\ and\ t = 1,\ -1 \end{cases} \quad (1)$$

where $C = (c_1,\ c_2,\ldots,c_n)$, $c_i$ is the color for elevation zone $i$, $n$ is the color amount in $C$, $R_k^2(y)$ is the goodness of fit in the $k$-degree polynomial fit $\sum_{m=0}^{k} a_m x^m, x = [1, 2, \ldots n]^T$, $f_L(t)$ is the correction function of luminance change, and *t* is the correction factor. Considering device independence and perceptual uniformity, all color computations here are performed in the CIELab space: *C.L* is the luminance component of *C*, and $C.ch_{ab}^* = \frac{\sqrt{a^2+b^2}}{\sqrt{L^2+a^2+b^2}}$ is the color chromaticity of *C*. *k* can be 1, 2, or 3, representing monotonous, dichromatic and polychromatic, respectively. When $t = 1$, $f_L(t)$ reflects "the higher, the darker" principle, and vice versa, -1 for "the higher, the lighter" principle.

**Aerial perspective**. The aerial perspective suggests a decrease in color contrast with increasing distance. Here, we understand it as the monotonicity of the color difference, denoted as $f_{ap}(C)$, which is as follows:

$$\begin{cases} f_{ap}(C) = \begin{cases} \frac{\sum \max(0, \Delta\Delta E_{ab}^*(c_i))}{\sum |\Delta\Delta E_{ab}^*(c_i)|} & \text{if } C \text{ follow aerial perspective} \\ 1 & \text{otherwisw} \end{cases} \\ \Delta\Delta E_{ab}^*(c_i) = \Delta E_{ab}^*(c_{i+1}, c_i) - \Delta E_{ab}^*(c_i, c_{i-1}) \end{cases} \quad (2)$$

where $\Delta E_{ab}^*$ is the Euclidean distance between two colors in CIELab.

*Color conventions*. We quantify the color convention as the degree of shifting from a 'standard' color for an elevation zone *i* as follows:

$$f_c(c_i) = \begin{cases} \min(1.0, \gamma / \Delta E_{ab}^*(c_i, c_i')) & \text{if } c_i \text{ follow a color convention} \\ 0 & \text{otherwise} \end{cases} \quad (3)$$

where $c_i'$ is the conventional color of elevation zone *i*, and $\gamma$ is the color distance threshold respecting the color convention. The total scores of respecting color conventions for the entire hypsometric tints *C* can be denoted as:

$$f_c(C) = \frac{1}{m} \sum_i f_c(c_i) \quad (4)$$

As above, we operationalize three subjective concerns, $F_s(C)$, by multiple $f_g$, $f_g$ and $f_c(C)$. As all three components are normalized, $F_s(C)$ ranges from 0 to 1.

### 3.3.2 Definition of similarity with aesthetic quality

We quantify the overall color similarity between an image and elevation as their color alignment to the identified dominant colors. Histograms are widely used to quantify color similarity between images. Since a terrain map's color proportion may be dramatically different from that of the reference image, overfitting of histogram matching is inevitable. Since key elevation colors all come from reference, we quality color similarity as the alignment of elevation colors to an image's dominant colors with consideration of their proportions. An image's color distribution is measured as $N$ dominant colors. For each dominant color, its proportion is accumulated over all the colors sharing the same region in the color grid. On the other hand, we anchor all elevation colors to dominant colors; the proportion of a dominant color is accumulated through all discrete colors in the graded color scheme and along the interpolation in the continuous color scheme. Since elevation colors may not be the dominant colors, we weigh its proportion according to its inverse distance to the dominant color. In this manner, all elevation colors are aligned to the dominant color with weighted proportion. The overall color similarity between the image and elevation colors can be measured as the degree of their alignment as follows:

$$\begin{cases} f_d(C) = \sum_{i=1}^{n} \left[ P_i^{image} * \frac{\min(P_i^{map}, P_i^{image})}{\max(P_i^{map}, P_i^{image})} \right] \\ P_i^{map} = \sum_j P_j^{map} * \min(1.0, \frac{\alpha}{\Delta E_{ab}^*(c_i^{image}, c_{i,j}^{map})}) \end{cases} \quad (5)$$

where $P_i^{image}$ is the proportion of the *i*-th dominant color in the reference image; $P_i^{map}$ is the weighted proportion of the *i*-th dominant color in the elevation colors; *j* is the *j*-th map color whose dominant color is *i*; $P_j^{map}$ is its proportion; and *α* is a color distance threshold.

Color similarity plays a double role in our model. First, we hope to use similar colors to provoke similar emotions. Second, we assume that the input reference is favorable to the users, and similar colors will therefore accommodate the user's color preference.

In addition to color preference, color harmony is one major aspect of aesthetic quality. We do not restrict that the reference should be perfect, but we hope to derive elevation colors with harmony. We therefore explicitly score color harmony. Here, we use the Kita and Miyata model (2016) to score the harmony of elevation colors, denoted *Harmony*(C). Overall similarity with aesthetic quality, $F_a$, is quantified as follows:

$$F_a(C) = f_s(C) * Harmony(C) \qquad (6)$$

We normalize the score on color harmony. Since color similarity is also normalized, the overall similarity with aesthetic quality also ranges from 0 to 1.

Since only elevation and area are needed, our model supports diverse terrain models, such as a DEM, TIN, and contours: elevation can be derived from grid, triangle, or contour; area can be calculated according to the amount of grid or triangle, or the coverage of contour fall into the elevation zone.

### 3.3.3 Problem solving

Based on the above scoring on subjective concerns and aesthetics quality, we formulize the color transfer problem as a dual-objective optimization problem, maximizing both $F_s(C)$ and $F_a(C)$. In addition to the above two objectives, constraints could be added optionally. For example, if graded elevation colors are desired, then a color discrimination constraint should be added to ensure that all colors are mutually distinguishable.

Color transfer from images to terrain maps is now formulized as a MAX-MAX problem. All objectives and constraints are clearly defined, and both Fs(C) and $F_a$(C) are normalized. All color variables are limited in a color grid, which can be solved using existing dual-objective optimization methods, such as an Artificial Bee Colony (ABC) (Omkar et al. 2011) or evolutionary algorithm (Deb 2014), with a series of color operations, such as local and global searching supported by a color grid. Note that dual-objective optimization often has a set of solutions in which one is not superior to another. These nondominated solution sets are mathematically called the *Pareto frontier*. Following the Wu et al. (2021) sampling strategy, we treat the midpoint of the *Pareto frontier* as the balanced solution in terms of the score on subjective concerns and similarity with aesthetic quality. The *Pareto frontier* can also be sampled to yield a series of satisfying solutions that may be favorable by experts. For example, to make seasonal terrain maps for a national park that exhibit an attractive valley for public tourists, a fantastic landscape painting from an artist is available by the designer as an inspiration. In this case, subjective concerns would be partly compromised. An optimum similarity with aesthetic quality may be desired to engage audiences.

## 4. Implementation and evaluation

## 4.1 Implementation with examples

We purposefully sampled four typical terrains to test our method, including a ridge, a canyon, a crater and a mountain (see Figure 4). All four terrains are gathered in DEM data format. Our method does not require color input at all. We color them using a standard color scheme from the 1962 International Map of the World for further comparison, and we extract the colors from Patterson and Jenny (2011).

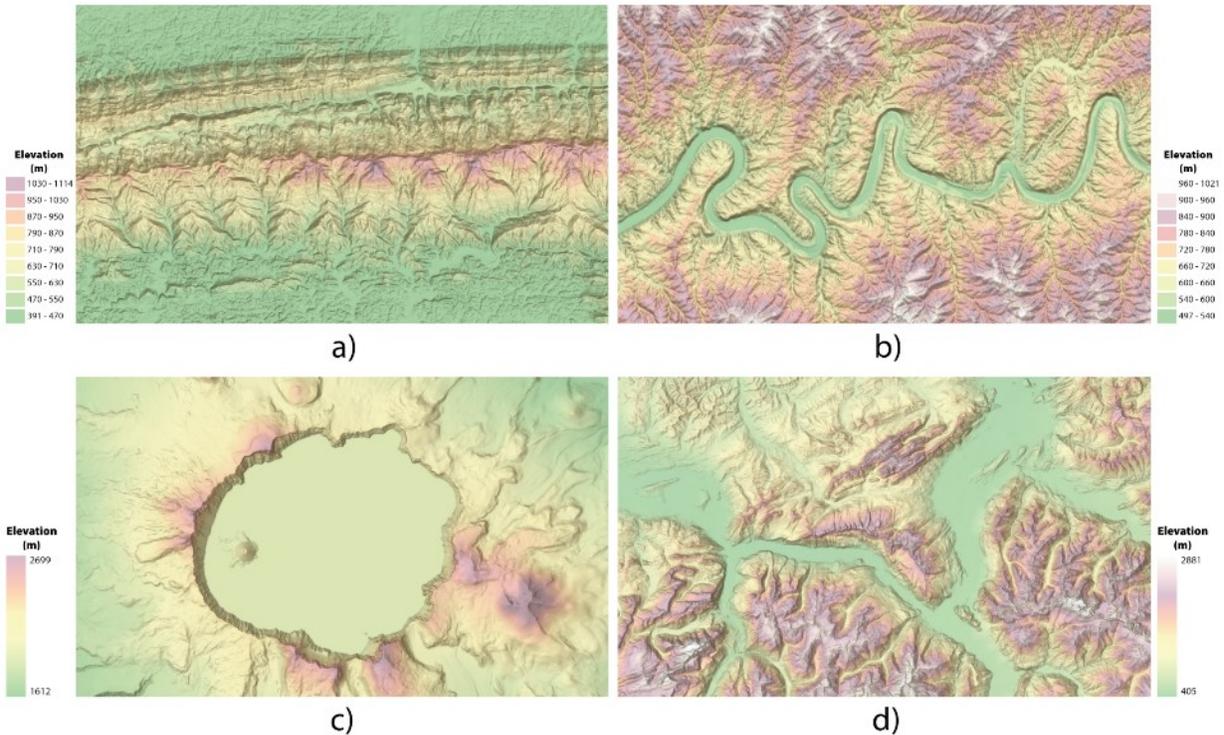

Figure 4. Four sample terrains. a) ridge, China (data source: NASA EARTHDATA (https://earthdata.nasa.gov), 12.5 meter resolution); b) canyon, China (data source: NASA EARTHDATA (https://earthdata.nasa.gov), 12.5 meter resolution); c) crater, the USA (data source: Sample Elevation Models (http://shadedrelief.com/SampleElevationModels/), 3.33 meter resolution); and d) mountain, Switzerland (data source: Sample Elevation Models (http://shadedrelief.com/SampleElevationModels/), 30 meter resolution).

Ten reference images are sampled via an image searching engine by using "landscape" as the key word. Five natural landscape photos and five paintings are collected, which are shown in Figure 5. Other types of images can also be candidates (see Section 5 for further discussion).

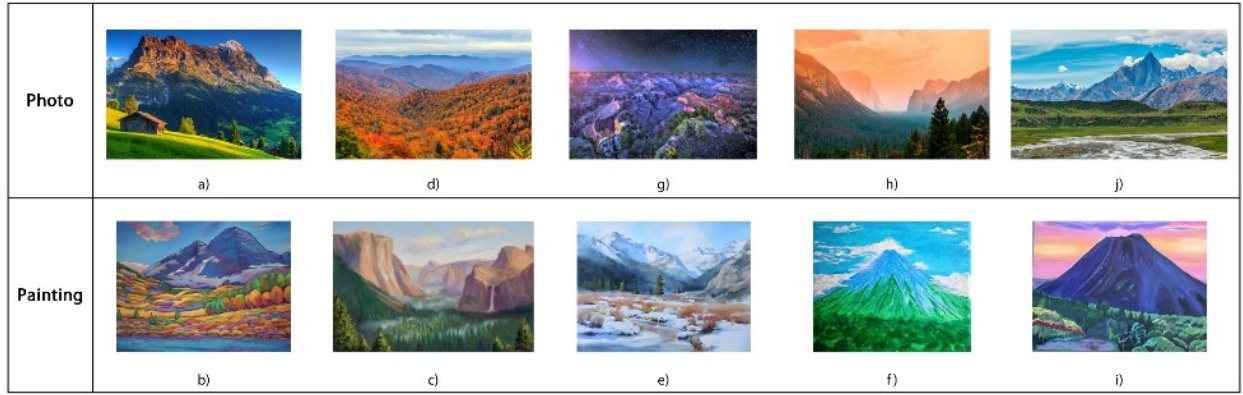

Figure 5. Ten sample reference images from the internet by using "landscape" as the image searching key word.

We implement our method in the C++ programming language; the source code and the test data can be found at https://doi.org/10.5281/zenodo.6509472. The involved parameters are assigned as follows across sample maps and reference images: three parameters related to color distance at formula 1 $t =1$, formula 3 $\gamma = 10$ and formula 5 $\alpha = 10$. Considering that the specific elevation range and location, the mountain and canyon are colored with multihued tints, the degree of polynomial for quantifying continuity ($k$) is set to 3. The crater and ridge are assigned dichromatic tints and $k = 2$. The results of the color transfer from reference a and reference b to the four terrains are shown in Figure 6.

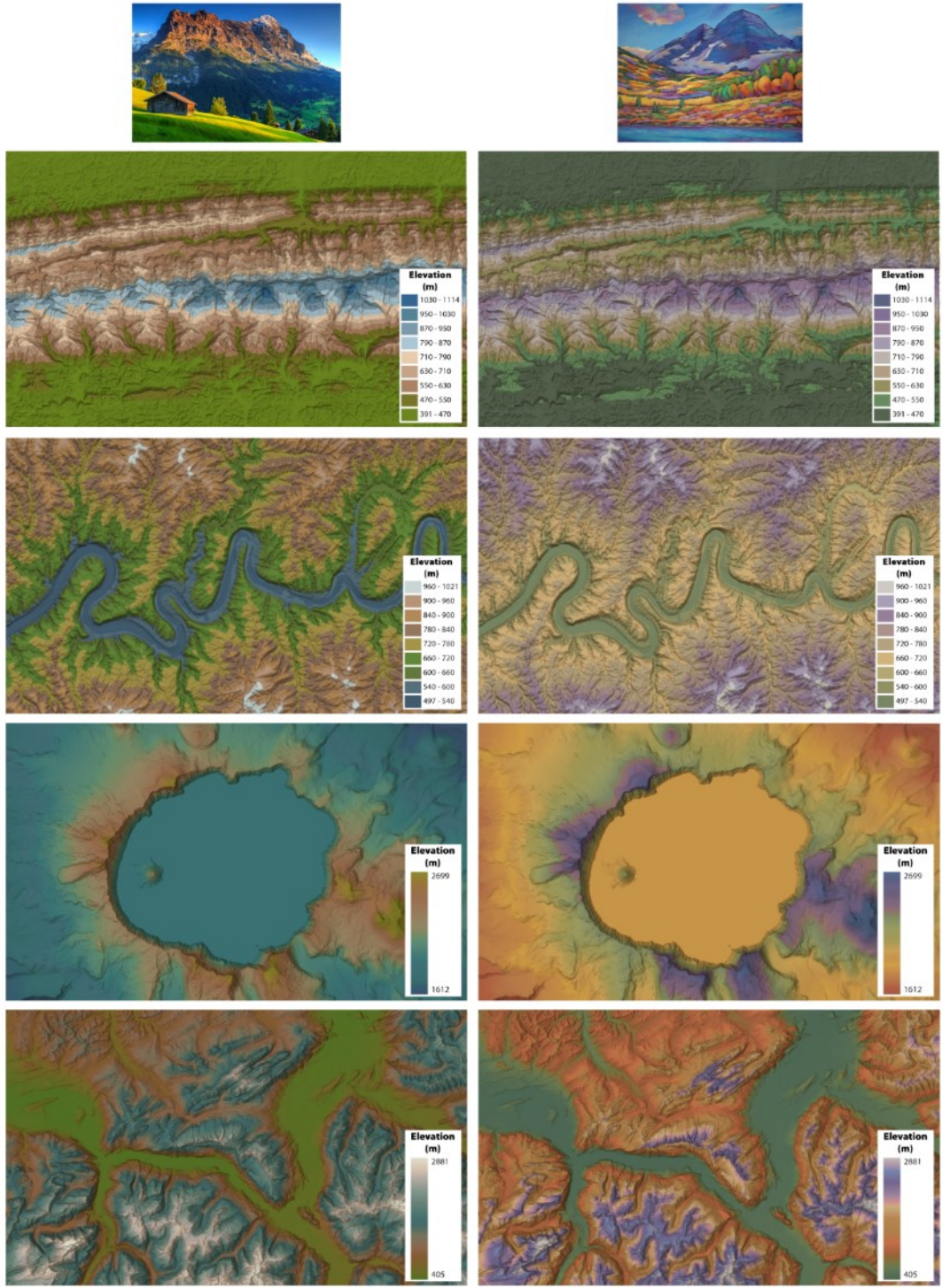

Figure 6. Results of color transfer from references a and b to the four terrains.

We use the canyon terrain and reference painting c to test our method's performance on supporting continuity. By setting *t* to 1 and -1, respectively, "the higher, the lighter "and "the higher, the darker" can be obtained (see Figure 7).

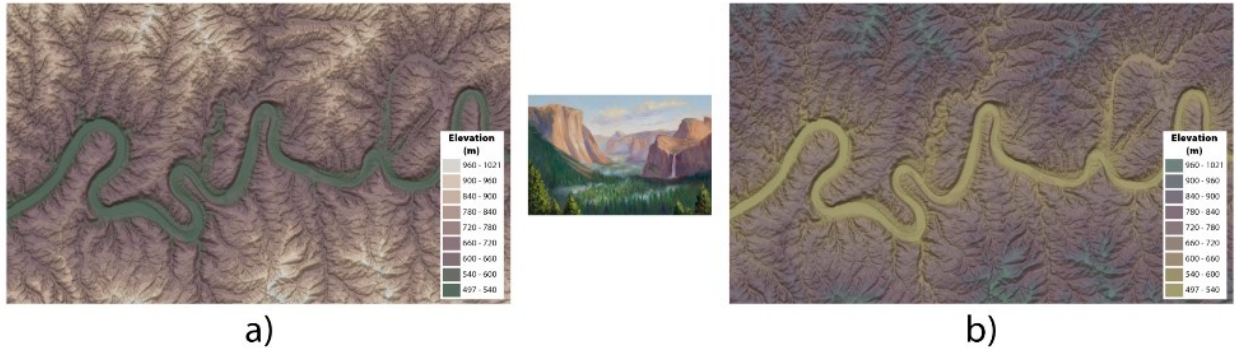

Figure 7. Performance on continuity (a) result of applying "the higher, the lighter" principle, *t* = 1 and (b) result of applying "the higher, the darker" principle, *t* = -1.

We use the ridge terrain and reference photo d to show the performance on supporting color conventions. The ridge is located in the subtropical seasonal climate zone in China, where grassland and forest grow. Therefore, we assign the lowland a green color. The results are shown in Figure 8a. For comparison, we also provide the result without color convention assignment (Figure 8b). Benefiting from our dual-objective optimization, both results are similar to the reference. While Figure 8b looks dustily, Figure 8a appears greenish, which is more coherent with the real landscape.

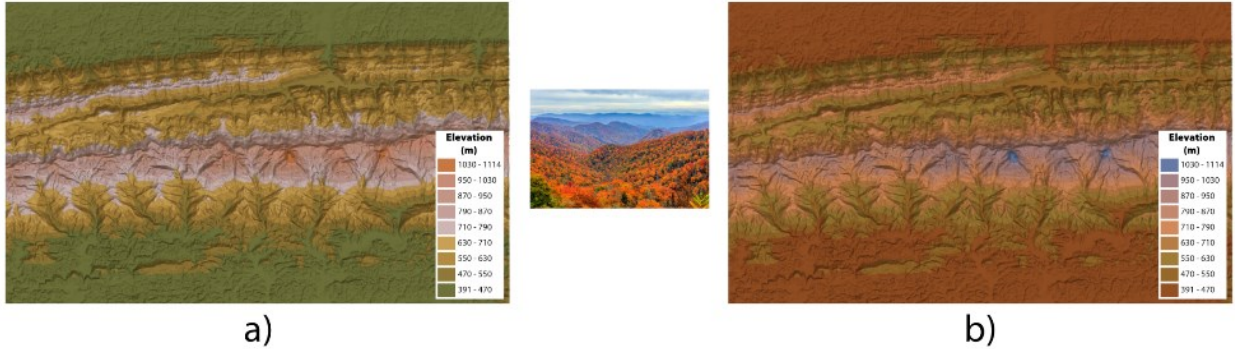

Figure 8. Results of applying a color convention. (a) Color relief with lowland associated with green color and (b) color relief without any color convention.

We also employ the mountain terrain and reference e to investigate our method's ability to make aerial perspective efforts. An aerial perspective effort can be made on color or hillshade. We make two gray hillshades on the mountain terrain, with and without an aerial perspective (see Figure 9a and b). We overlap our tints with the aerial perspective in Figure 9a and b for comparison (See Figure 9c and d). This shows that hypsometric tints with areal perspective effort from our method can help to enhance the "the farther, the less contrast" principle.

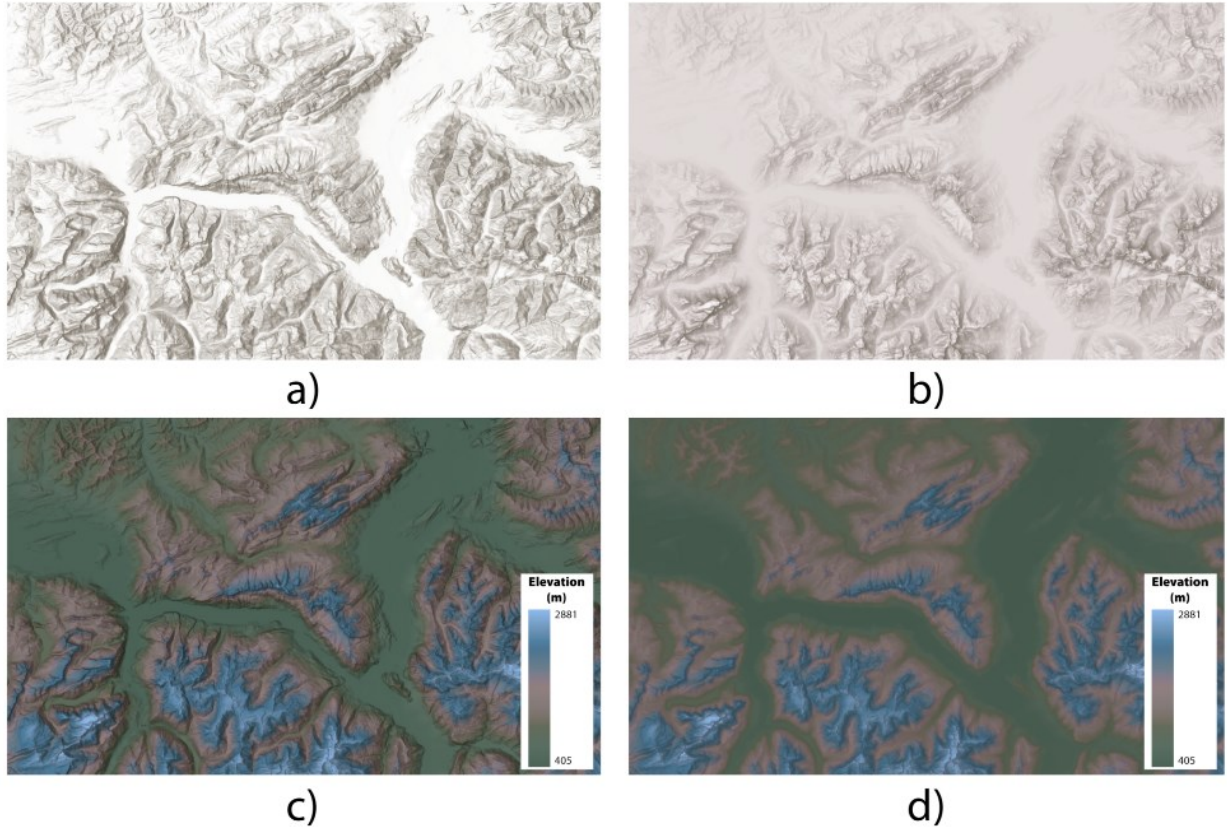

Figure 9. Comparison of aerial perspectives. (a) gray hillshade without aerial perspective; (b) gray hillshade with aerial perspective; (c) colors with aerial perspective on the gray hillshade without aerial perspective; and (d) colors with aerial perspective on the gray hillshade with aerial perspective.

Finally, we use the crater terrain and reference painting f to investigate the performance on diverse terrain models. We derive the TIN and contours from the row DEM data and then identify elevation zones and calculate the corresponding area from them. The results of shaded TIN and illuminated contour are shown in Figure 10a and b. For hillshade from the DEM, multidirectional hillshade is common, but feature-enhanced hillshade (such as aspect and curvature) is also widely used in practice. We overlap our tints with a multidirectional hillshade (see Figure 10c) and a curvature-enhanced hillshade (see Figure 10d). They share a visual appearance similar to that of the reference.

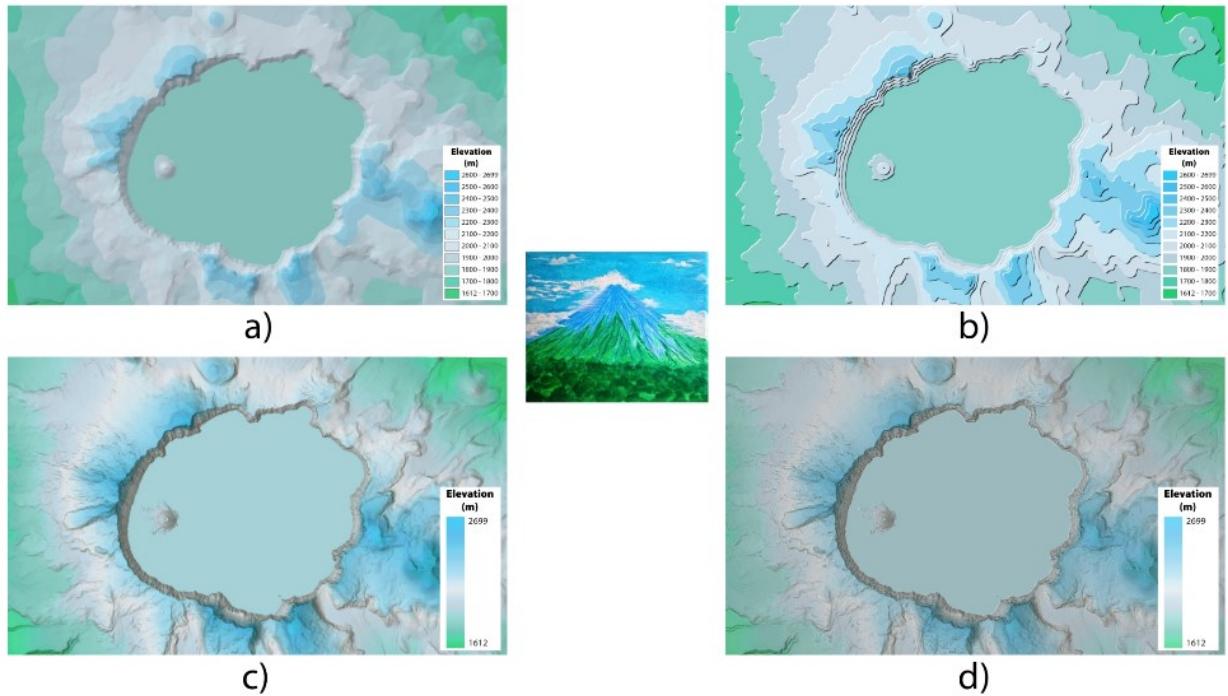

Figure 10. Resulting color tints on diverse terrain models. (a) TIN; (b) illuminated contours; (c) multidirectional hillshade; and (d) curvature-enhanced hillshade.

## 4.2 Evaluation with standard color tints

To the best of our knowledge, there is no counterpart method that can adaptively transfer color from images to terrains. We therefore compared the output color scheme from our method with the standard scheme in Figure 5. We design three tasks to examine their performance in terms of efficiency. We used one questionnaire for aesthetic quality and one questionnaire for affective response. We describe the materials, the participants, the procedure and the results as follows.

### 4.2.1 Materials

We select four reference images for each test map. As shown in Figure 6, we use photo a and painting b as two common references. We further use photos d and g for map A, painting c and photo h for Map B, paintings f and i for Map C, and painting e and photo j for Map D. In total, we make four styled maps for each test map. The first three are shown in Figures 5, 6, 7 and 8. The fourth for each test map can be found in Figure 11. To avoid color distortion between displaying devices, all materials are displayed on a desktop computer with a screen resolution of 1920 × 1080 pixels in a distraction-free room.

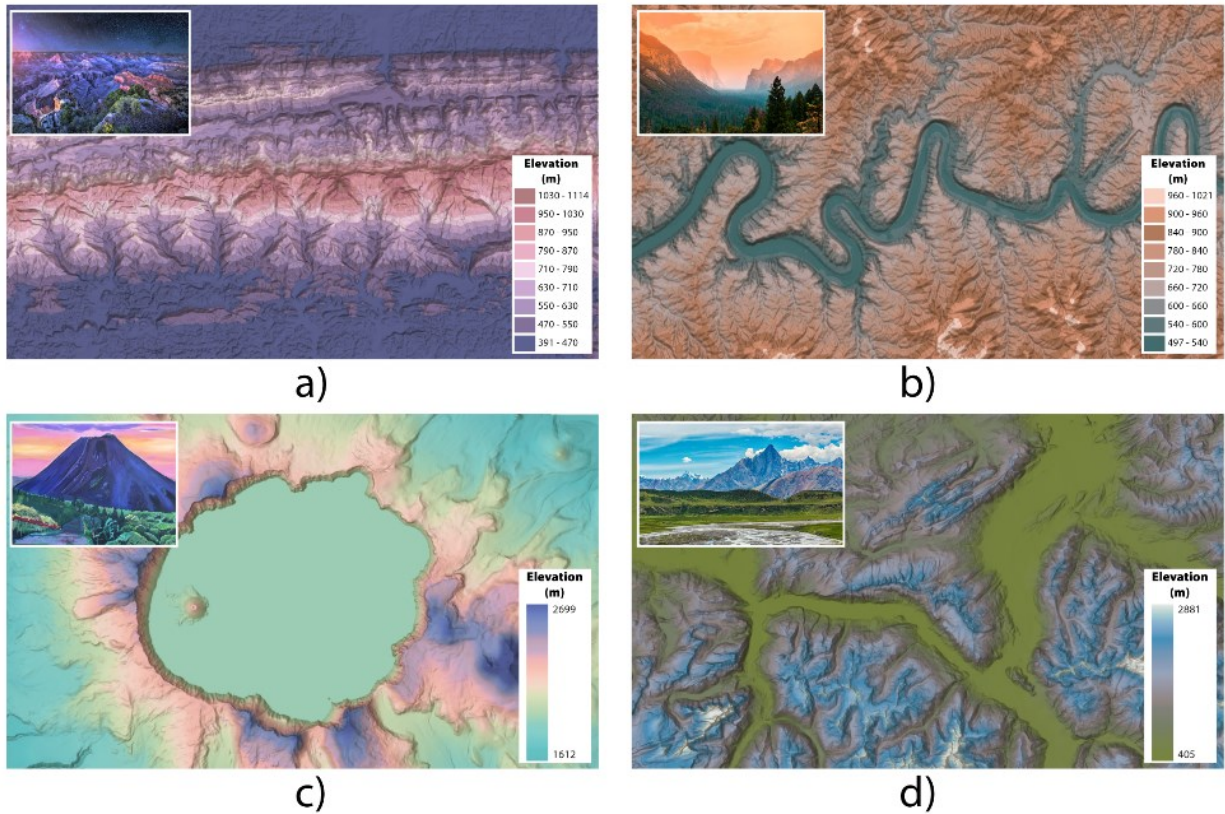

Figure 11. Four styled maps using references g, h, i and j respectively.

## 4.2.2 Procedure

We design three tasks to evaluate the cartographic quality: *identify*, *locate* and *compare* (see Figure 12). For the *identify* task, we asked participants to *identify* the elevation zone of three test points; nice elevation zones are given in the legend. For the *locate* task, we asked participants to point out which side of the highlighted region (i.e., east, south, west or north) is the lowland. For the *compare* task, participants were asked to *compare* the elevations of three points. All testing points and regions are randomly sampled across the whole terrain. We use the same testing point and areas for both comparisons. All answers are assessed as a binary correct/incorrect. Then, the accuracy of the rate of correct answers can be calculated.

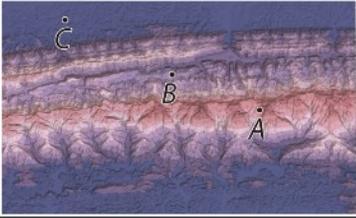

Figure 12. Three tasks used to compare elevation colors from styled map with standard elevation colors.

We designed a questionnaire to evaluate esthetic quality:

**Question 1:** At which level do you agree that the color relief is visually pleasing?

Seven-point scales are used to record participant feedback, from 1 (totally disagree) to 7 (strongly agree).

We select references g, h, i and j and their corresponding styled map as the four test maps to investigate affective response. We designed a questionnaire to compare their affective response. Here, we use the hourglass emotion model, which distinguishes four dimensions of emotion: pleasantness, attention, aptitude and sensitivity (Cambria et al. 2012). Before scoring, we explained these four dimensions to all participants with the four emotion pairs derived from (Cambria et al. 2012): sadness-joy for pleasantness; distraction-interest for attention; disgust-trust for aptitude; and anger-fear for sensitivity. For each dimension, three levels of response are designed in this evaluation: positive (1), neural (0) and negative (-1).

We then asked the following:

**Question 2:** At which level do your affective response fall for the given material?

Four dimensions are scored separately. To determine whether the emotion embedded in the reference image is transferred or not, we also invite participants to score their response on the reference, but we do not inform them which color relief is transferred from which reference.

Furthermore, we believe that in addition to color, many other factors will impact an observer's affective response, so we also asked them the following question after they scored the materials:

**Question 3:** What factors impact your affective response when scoring the styled maps and references?

In total, 54 participants were invited to conduct the tasks. All participants were students at the authors' college. None of the participants possessed an eye disease or color vision deficiency. Each participant read four terrain maps: two styled maps from different terrains (randomly selected for each participant) and two other terrain maps in standard color scheme. We recorded the accuracy and the response time for the three tasks on each map.

Twenty-six participants were invited to complete the two questionnaires. We invited them to score the aesthetic quality first and then affective response. To avoid being impacted by the last review, we show a blank page to them before reading any piece of material.

### 4.2.3 Results

The statistical results of the accuracy are shown in Table 1. It shows that there is a slight difference in accuracy between our tints and the standard tints, suggesting that our tints can work as well as the standard. Note that 2 cases (out of 12) pass the significance test ($p$-value < 0.05), suggesting that our tints are occasionally better than the standard.

Table 1. Independent sample test for accuracy on the tasks.

|  | Map | Differences (styled map – standard map) | | $t$ | df | $p$-value |
|---|---|---|---|---|---|---|
|  |  | M | SE |  |  |  |
| *Identify* | A | 0.024 | 0.024 | 1.000 | 52 | 0.322 |
|  | B | -0.012 | 0.012 | -1.000 | 52 | 0.322 |
|  | C | 0.086 | 0.042 | 2.054 | 52 | 0.045* |
|  | D | 0.037 | 0.036 | 1.018 | 52 | 0.314 |
| *Locate* | A | 0.148 | 0.070 | 2.126 | 52 | 0.038* |
|  | B | 0.000 | 0.073 | 0.000 | 52 | 1.000 |
|  | C | 0.000 | 0.052 | 0.000 | 52 | 1.000 |
|  | D | 0.111 | 0.102 | 1.087 | 52 | 0.282 |
| *Compare* | A | 0.074 | 0.051 | 1.442 | 52 | 0.155 |
|  | B | 0.000 | 0.052 | 0.000 | 52 | 1.000 |
|  | C | -0.037 | 0.037 | -1.000 | 52 | 0.322 |
|  | D | 0.037 | 0.037 | 1.000 | 52 | 0.322 |

$M$ = mean, SE = standard error.

*Difference is significant ($p < 0.05$).

The statistical results of the time taken are shown in Table 2. Similar to the accuracy performance, Table 2 suggests that our tints and the standard tints are at the same effectiveness level. Our method performed better in 2 cases (out of 12).

Table 2. Independent sample test for the time taken in seconds to complete the tasks.

|  | Map | Differences (styled map – standard map) M | SE | t | df | p-value |
|---|---|---|---|---|---|---|
| Identify | A | 3.407 | 2.772 | 1.229 | 52 | 0.224 |
|  | B | 0.892 | 1.818 | 0.491 | 52 | 0.626 |
|  | C | 0.615 | 1.994 | 0.308 | 52 | 0.759 |
|  | D | -0.792 | 3.198 | -0.248 | 52 | 0.805 |
| Locate | A | -3.981 | 1.871 | -2.128 | 52 | 0.038* |
|  | B | 0.030 | 1.505 | 0.020 | 52 | 0.984 |
|  | C | 0.267 | 0.778 | 0.343 | 52 | 0.733 |
|  | D | -0.655 | 2.331 | -0.281 | 52 | 0.780 |
| Compare | A | -0.004 | 1.499 | -0.002 | 52 | 0.998 |
|  | B | -2.362 | 1.158 | -2.040 | 52 | 0.046* |
|  | C | -0.163 | 1.300 | -0.125 | 52 | 0.901 |
|  | D | 3.396 | 4.269 | 0.796 | 52 | 0.430 |

$M$ = mean, SE = standard error.

*Difference is significant ($p < 0.05$).

Generally, the above preliminary results show that tints from our method are not inferior to standard hypsometric tints. Benefitting from the high quality of the references, our tints are partly better than standard hypsometric tints.

The results of the evaluation of aesthetic quality are shown in Table 3. The results suggest that the styled maps are more aesthetically favorable: 10 tests (out of 13) are statistically significant ($p$-value < 0.05).

Table 3. Paired samples test for the scores on aesthetics.

| Map | Reference | Paired differences (styled map – standard map) M | SD | t | df | p-value |
|---|---|---|---|---|---|---|
| A | a | 1.077 | 1.412 | 3.889 | 25 | 0.001* |
|   | b | 0.808 | 1.415 | 2.911 | 25 | 0.007* |
|   | c | 0.962 | 1.536 | 3.193 | 25 | 0.004* |
|   | e | 0.115 | 1.395 | 0.422 | 25 | 0.677 |
| B | a | 1.115 | 1.583 | 3.593 | 25 | 0.001* |
|   | b | 0.538 | 1.529 | 1.795 | 25 | 0.085 |
|   | d | 0.885 | 1.423 | 3.169 | 25 | 0.004* |
|   | g | 0.846 | 1.317 | 3.275 | 25 | 0.003* |
| C | a | 1.077 | 1.623 | 3.384 | 25 | 0.002* |
|   | b | 0.692 | 1.619 | 2.180 | 25 | 0.039* |
|   | f | 1.038 | 1.907 | 2.776 | 25 | 0.010* |
|   | h | 1.000 | 1.789 | 2.850 | 25 | 0.009* |
| D | a | 0.692 | 1.543 | 2.287 | 25 | 0.031* |
|   | b | 0.423 | 1.554 | 1.389 | 25 | 0.177 |
|   | i | 0.923 | 1.468 | 3.207 | 25 | 0.004* |
|   | j | 0.885 | 1.558 | 2.896 | 25 | 0.008* |

$M$ = mean, SD = standard deviation.

*Difference is significant ($p < 0.05$).

The affective response varies across terrains and references. The affective response on styled maps vs. the standard map and styled maps vs. the reference (using references b, c, f and j) are shown in Figure 13. It shows that 10 pairs of emotions (out of 16) are significantly different ($p$-value < 0.05) when comparing styled maps with the standard maps, suggesting emotions carried by the maps are shifted after styling; 15 pairs of emotions (out of 16) are very similar (insignificant difference) when comparing the styled maps with the reference images, suggesting emotion are also transfer well from reference to styled maps. When referring to references d, h, i and a, 7 pairs of emotions (out of 16) are significantly different, and 7 pairs of emotions (out of 16) are similar (see Figure 14), indicating that emotions are partially transferred.

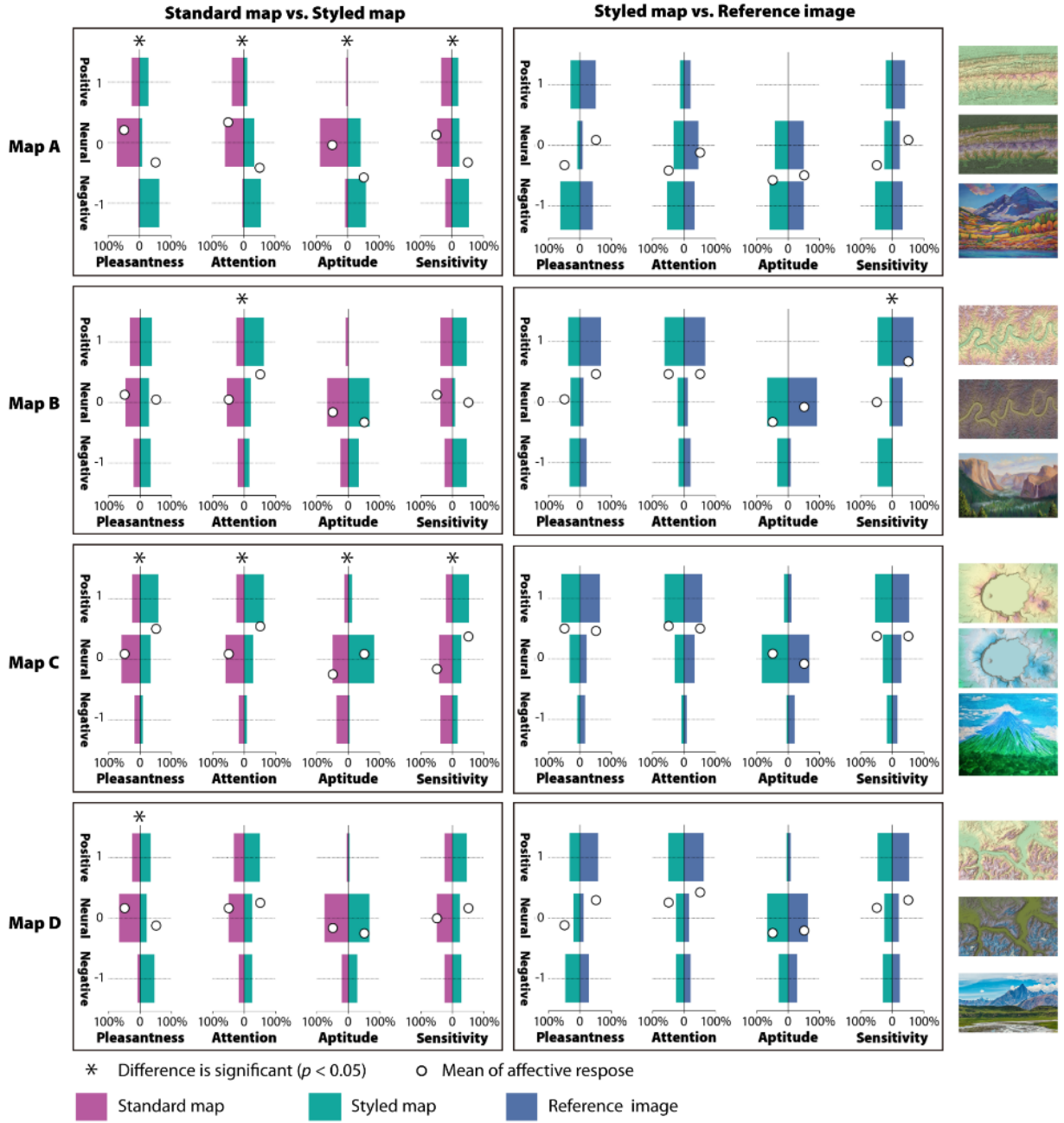

Figure 13. Comparison of affective response to the standard maps, the styled maps and the reference images, using references b, c, f and j respectively.

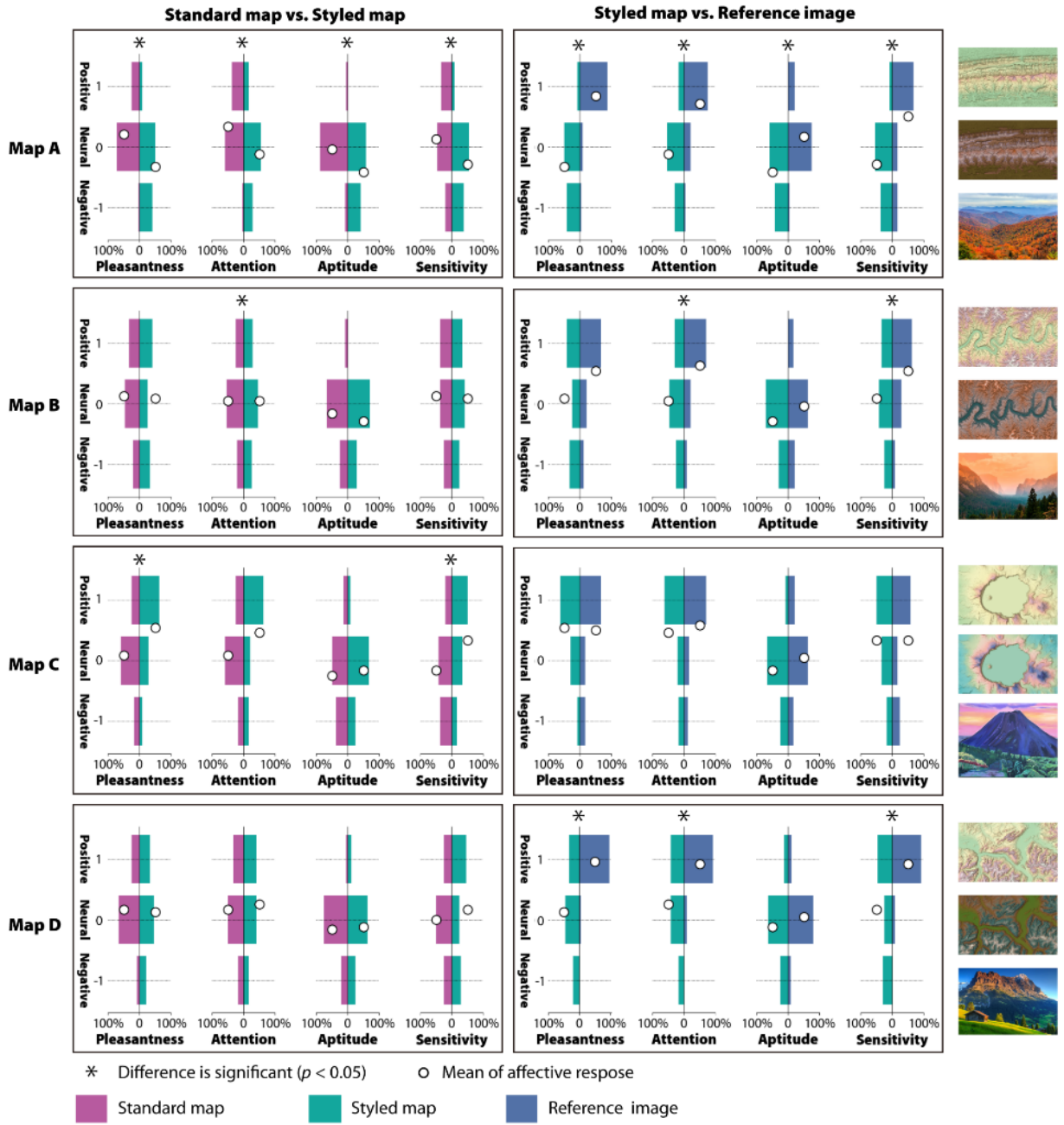

Figure 14. Comparison of affective response to the standard maps, the styled maps and the reference images, using references d, h, i and a respectively.

We deeply investigate what factors impact a map reader's emotions. We pool all feedback from **Question 3** into Figure 15. It shows that color plays a critical role, but content, texture, form, background and others (e.g., viewpoint) are also important. Importantly, those facts impact affective response at different levels for styled maps and reference images. For instance, many participants mentioned the background and surroundings of the reference image, but very few participants mentioned that fact on styled maps. Note that since all data are at high resolution, topological details (e.g., aspect and slope) (generally called texture here to avoid using too much expertise for the general participants) are presented; major participants agree that texture significantly impacts their affective response on styled maps, but texture influences their affective response on reference much less. When transferring color from reference to terrain, if content, form, texture and background are all matched well, then emotion can be transferred faithfully. However, it is too restricted in practice, and it will also dramatically narrow the scope of color inspiration. We therefore claim that our method can conditionally transfer emotions from reference images to terrain maps.

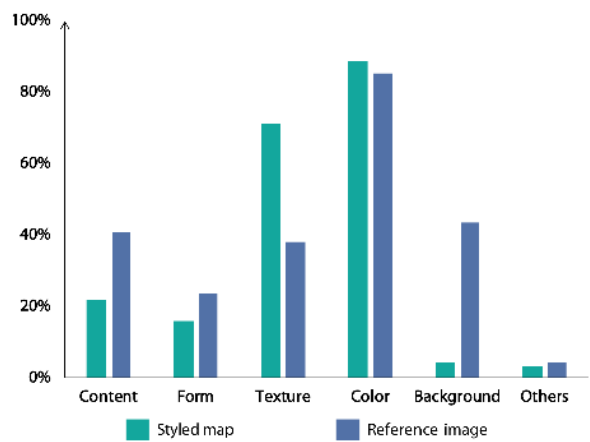

Figure 15. Statistical results of emotional factors for both styled maps and reference images.

## 5. Discussion

Although our method shows advantages in adaptive image-to-terrain color transfer with high aesthetic quality, it also has several limitations for further consideration:

First, **semantic matching**. As mentioned in Section 2.2, there are a series of methods that transfer color by keeping the semantic constant. In this study, we do not model semantic matching to relax the constraint that the reference should include elements about terrain. We select all references using "landscape" as the key word. However, other types of images, such as portraits and still life paintings, can also be used as references. As shown in Figure 16, we apply van Gogh's two paintings on the carter terrain. Benefitting from his expressive colors, the styled maps all look stylish. In this paper, semantic-color consonants are considered color conventions; semantic matching will be addressed in the future.

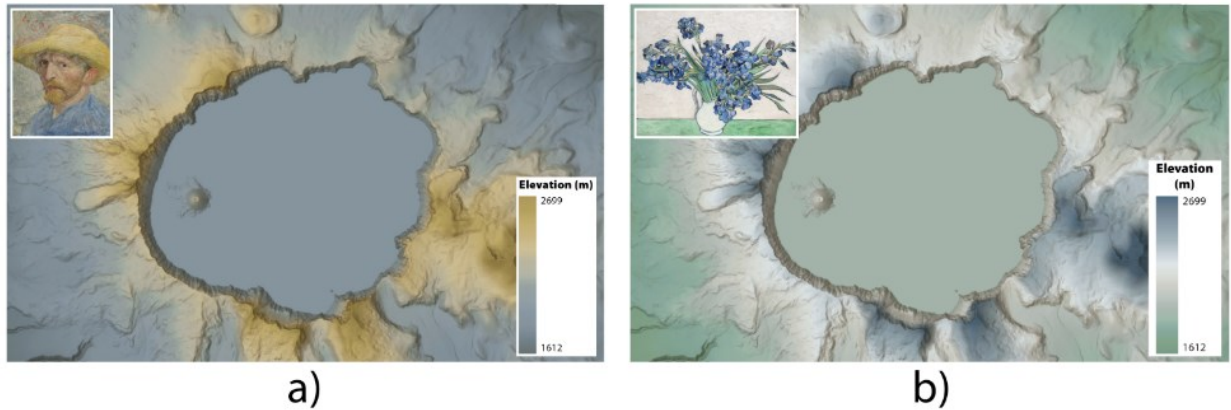

Figure 16. Results of transfer color from van Gogh's two paintings: (a) self-portrait; and (b) irises.

The second is **adaptability**. As discussed in the Introduction, naturalism and symbolism should be balanced when crafting expressive hypsometric tints. Their compromise is relayed to the diversity of terrain. In our model, we quantify color convention as color shifting from a given representative color, which is initialized using a standard scheme. However, the representative color for elevation zones may vary across space (e.g., climate zones). Furthermore, we apply multihued tints on the mountain and canyon, dichromatic tints to the crater and ridge, and proper hue ranges also vary across space. To adapt to diverse terrains, these parameters should be extended in the future.

Third, **reference suggestion**. We show that our method can transfer arbitrary images to diverse terrains. However, this does not mean that all images are suitable for good color inspiration. As shown in Tables 1 and 2, the cartographic quality also varies on different terrains. Then, a question is naturally raised: how can a proper reference be chosen for a given terrain? Many other map-use contexts should also be considered, such as the audience and activities (e.g., way finding). These considerations will be addressed in the future.

## 6. Conclusion

In this paper, we argue that terrain mapping not only communicates how high or how steep a landscape is but also includes how we feel at a very place. Both subjective concerns and aesthetic quality should be taken into account, as should emotion. Involving much expertise in cartography, topography and color science, crafting expressive and affective hypsometric tints for terrain mapping is challenging for both nonexperts and experts. In this paper, we present an image-to-terrain color transfer method that can transfer color from arbitrary images to diverse terrain models. Our contributions can be highlighted as follows:

First, we present a new image color organization method. We use the SOM algorithm and hierarchal clustering technique to organize discrete, irregular image colors into a continuous, regular color grid. It is able to facilitate a series of color operations, such as local color adjustment and global color exploration, categorical color selection and sequential color interpolation. It adapts to diverse images.

Second, we mathematically treat image-to-terrain color transfer as a dual-objective optimization problem. We operationalized a series of subjective concerns of terrain mapping, such as "the lower, the higher" principle, color conventions, and aerial perspective. We also define color similarity between the image and terrain map with aesthetic quality. The evaluations show that the hypsometric tints from the proposed method can work as well as the standard tints, and our tints are more visually favorable. We also showcase that our method can conditionally transfer emotion.

Our method can potentially facilitate making effective, beautiful and affective terrain maps. A series of future work still need to be addressed, such as semantic matching and reference suggestions; extensions are also anticipated to be adaptive to diverse terrains.